\documentclass[letterpaper, 10 pt, conference]{ieeeconf}
\IEEEoverridecommandlockouts
\newcommand{\comment}[1]{}

\usepackage{atbegshi,picture}
\AtBeginShipoutNext{\AtBeginShipoutUpperLeft{%
  \put(\dimexpr\paperwidth-1cm\relax,-1.5cm){\makebox[0pt][r]{Accepted manuscript IEEE/EMBS Neural Engineering (NER) 2021}}%
}}

\usepackage[utf8]{inputenc}
\usepackage{xcolor, pdflscape, pdfpages, siunitx}
\usepackage{amssymb}
\usepackage{amsmath}
\usepackage{graphicx, subfigure}
\usepackage{caption}
\usepackage[noadjust]{cite}

\title{\LARGE \bf Neuromechanics-based Deep Reinforcement Learning of Neurostimulation Control in FES cycling}
\author{
Nat Wannawas$^1$, Mahendran Subramanian$^{2}$, and A. Aldo Faisal$^{1,2}$
\thanks{Brain \& Behaviour Lab: $^1$Dept. of Bioengineering , $^2$Dept. of Computing, Imperial College London, London SW7 2AZ, UK.
We acknowledge funding from Royal Thai Government Scholarship and a UKRI Turing AI Fellowship to AAF.
}
}

\begin{document}

\maketitle

\begin{abstract}
    Functional Electrical Stimulation (FES) can restore motion to a paralysed's person muscles. Yet, control stimulating many muscles to restore the practical function of entire limbs is an unsolved problem. Current neurostimulation engineering still relies on 20th Century control approaches and correspondingly shows only modest results that require daily tinkering to operate at all. Here, we present our state-of-the-art Deep Reinforcement Learning developed for real-time adaptive neurostimulation of paralysed legs for FES cycling. Core to our approach is the integration of a personalised neuromechanical component into our reinforcement learning (RL) framework that allows us to train the model efficiently--without demanding extended training sessions with the patient and working out-of-the-box. Our neuromechanical component includes merges musculoskeletal models of muscle/tendon function and a multi-state model of muscle fatigue, to render the neurostimulation responsive to a paraplegic's cyclist instantaneous muscle capacity. Our RL approach outperforms PID and Fuzzy Logic controllers in accuracy and performance. Crucially, our system learned to stimulate a cyclist's legs from ramping up speed at the start to maintaining a high cadence in steady-state racing as the muscles fatigue. A part of our RL neurostimulation system has been successfully deployed at the Cybathlon 2020 bionic Olympics in the FES discipline with our paraplegic cyclist winning the Silver medal among 9 competing teams.
\end{abstract}

\section{Introduction}
Functional Electrical Stimulation (FES) is a technique that induces involuntary contraction of muscles by stimulating the muscles using low-energy electrical signals. FES can induce the contraction of paralysed muscles, thereby allowing paralysed patients to move their limbs. FES-induced physical exercises such as FES-Cycling, one of the most widely performed FES exercises, in paralysed patients can help prevent adverse health effects such as muscle atrophy and osteoporosis \cite{Ferrante2008}.

\begin{figure}[h]
    \begin{center}
    \includegraphics[width=\columnwidth]{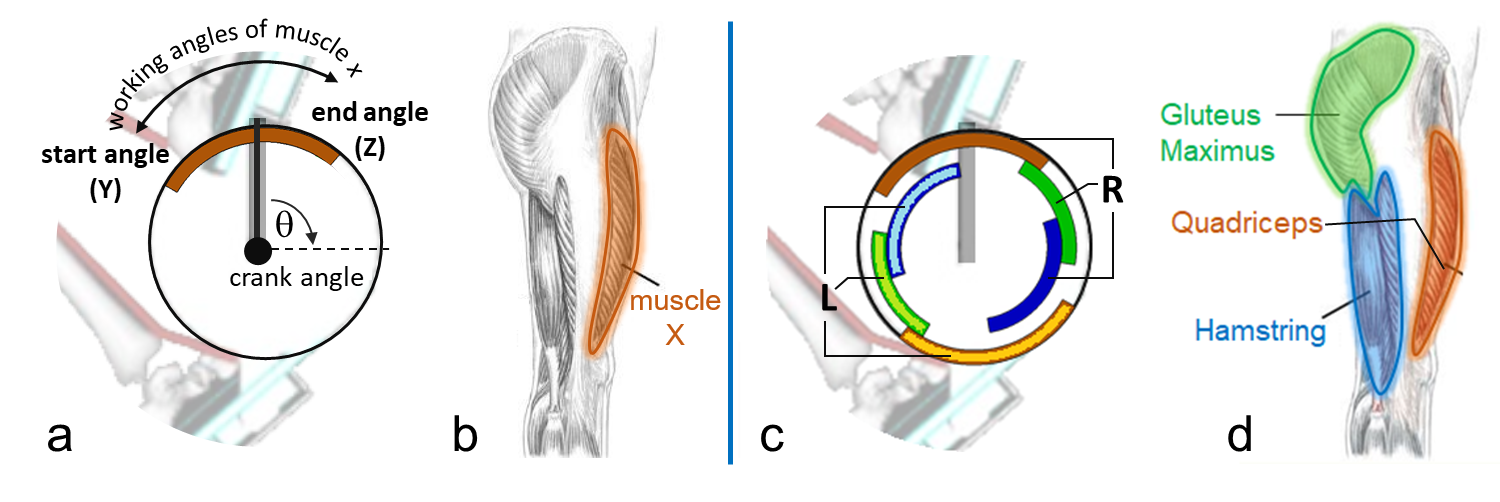}
    \captionsetup{margin=2mm, belowskip=-6mm, font=small}
    \caption{Illustration of the cyclical stimulation pattern (a) and corresponding muscle (b). (c,d) shows our multi-muscle.
    \label{fig:DesProfile}}
    \end{center}
\end{figure}

A conventional control method of FES-cycling is as follow. The cycling movement is induced by applying a pattern of FES to leg muscles. The stimulation of each muscle depends primarily on the angle of the cycling crank (Fig. \ref{fig:DesProfile} a and b). The stimulation of all involved muscles is called a stimulation pattern or ON/OFF angles (Fig. \ref{fig:DesProfile} c and d). The cycling cadence is controlled by varying the stimulation intensity \cite{DeSousa2016}, which is either manually controlled by a user or a control method such as PID and Fuzzy Logic Control.

The challenges of controlling FES in FES-cycling are as follows. Firstly, the optimal stimulation patterns, which are different across individuals, has to be optimized manually \cite{Dunkelberger2020}. This manual optimization is time-consuming and requires technical knowledge. Secondly, the controller has to deal with non-linear and non-stationary responses of muscles due to, for example, muscle fatigue.

This work approaches the challenges of controlling FES in FES-cycling using a data-driven method: Reinforcement Learning (RL) \cite{Sutton1998}, the machine learning framework where we learn how to control a system through interacting with it. Reinforcement learning for FES control has been examined in several studies. A simulation study by Thomas et al. \cite{Thomas2008} presented the ability RL in controlling FES for arm movements. Febbo et al. \cite{DiFebbo2018} successfully applied RL to control elbow angle using FES in real-world. These studies presented the potential of RL in FES applications.

Here, we study the application of RL to neurostimulation in a model system of FES cycling. We create RL agents that learn to stimulate the muscles to induce the cycling movement as well as to control the cadence. We train the agents on a musculoskeletal model in Opensim software. We modify the model to include muscular fatigue. We then compare the control performances of the RL agents with those of the PID and the Fuzzy Logic controllers. Our study presents the potential of RL in area of neurostimulation, specifically FES, in three aspects. Firstly, we demonstrate that RL is a feasible technique that can automatically discover a personalized stimulation pattern. This automatic discovery can tremendously reduce the amount of time and effort involved in setting up the cycling system. Secondly, we demonstrate that the RL controller outperforms conventional controllers in several aspects. Lastly, we demonstrate the ability of RL in dealing with muscular fatigue.

\section{Methods}
To demonstrate the potential of RL on FES-cycling under advancing muscular fatigue, we first describe the RL algorithm and the architecture of choice. Next, we formulate the RL training and briefly describe the PID and the FL controllers to which we compare. We also include information about the musculoskeletal and the muscular fatigue model.
\subsection{RL algorithm and training}
This subsection describes the RL algorithms of choice and the architecture, followed by the formulation of RL problem and the training settings.

Our Deep RL algorithm is based on the actor-critic with deep deterministic policy gradient (DDPG) \cite{Lillicrap2015}, a model-free RL algorithm suitable for control tasks in continuous state-action spaces. Fully connected neural networks are used to approximate the actor and the critic. Both actor's and critic's networks consist of 2 hidden layers with 250 neurons and the ReLU activation function. The actor's network has 6 neurons with the Sigmoid activation function in the output layer. Target networks with soft target update are used to improve the stability of the training \cite{Lillicrap2015}. We create two RL agents: Starter and Tracker, which together form the RL controller. The Starter is responsible for initiating the cycling movement. It brings the cadence up to 5 rad/s. After that, the Tracker takes over the system to track the desired cadences.

The overview task of the RL agents here is to apply the stimulation on 6 leg muscles (3 muscles on each leg) to induce cycling movement at desired cadences. The task is formulated based on the Markov Decision Process (MDP) briefly described as follows. At each time-step, the agents perceive a state vector $\mathbf{s}\in\mathbb{R}^9$ ($\mathbf{s} \in\mathbb{R}^8$ for the Starter)  consisting of crank angle $\theta$, cadence $\dot{\theta}$, desired cadence $\dot{\theta}_d$ (only Tracker), and 6 elements of fatigue factor $f\in[0,1]$. The detail of the fatigue factor is described in muscle fatigue model section. The agent then outputs a 6x1 action vector whose elements $s_i\in[0,1]$ are the stimulation for each involved muscle. The goal of the agents is to learn the stimulation policies that maximize numerical rewards. The rewards for the Starter and the Tracker are computed as Eq. \ref{eq:rewardStarter} and Eq. \ref{eq:rewardTracker}, respectively.

\begin{equation}
\label{eq:rewardStarter}
    r_{{S}}(t) = 
\begin{cases}
    \dot{\theta}_t - \frac{1}{6}\sum_{i=1}^6 s_{i,t}^2  &,  \dot{\theta_{t}} < 5\:\mbox{rad}/s \\
    100 - \frac{1}{6}\sum_{i=1}^{6}s_{i,t}^2 &, \mbox{otherwise}
\end{cases}
\end{equation}

\begin{equation}
\label{eq:rewardTracker}
    r_{{T}}(t) =
    -|\dot{\theta}_t - \dot{\theta}_{d,t}| - \frac{1}{6}\sum_{i=1}^6 s_{i,t}^2
\end{equation}
where $r_S(t)$ and $r_T(t)$ are the immediate rewards for the Starter and the Tracker at time $t$; and $\dot{\theta}_t$ is the cadence at time $t$ in $\mbox{rad}/s$; $\dot{\theta}_{d,t}$ is the desired cadence at time $t$ in $\mbox{rad}/s$; $s_{i,t}$ is the stimulation on muscle $i$ at time $t$,

The intuition behind both reward functions is explained as follows. The term $\sum_{i=1}^{6}s_{i,t}^2$ penalizes the stimulation which aims to minimize the effort. Next, the reward for the Starter (Eq. \ref{eq:rewardStarter}) is divided into two cases. The first case is when the cadence is less than the desired value (5 rad/s). Here, the term $\dot{\theta}_t$ encourages the agent to induce higher cadence. The second case triggers when the cadence reaches the desired value. Here, the agent receives a big reward of 100, and the training episode terminates thereafter. The reward function of the Tracker (Eq. \ref{eq:rewardTracker}) penalizes the cadence error which is the absolute difference between an actual and a desired cadences.

The training is episodic with the maximum number of steps of 100 and 100 ms time-step size. The desired cadence for the Tracker is randomly selected at the beginning of an episode and remains constant throughout the episode. An episode starts at a random crank angle. In practice, the Starter is trained first. The trained Starter is then used to initiate the cycling of the Tracker's training.

\subsection{Conventional Controllers}
In this work, we compare the performance of RL controller to those of two conventional controllers used in FES to date: the PID and the Fuzzy Logic controllers. The configurations of both controllers, which were designed to work on this Opensim model, were adopted from \cite{DeSousa2016}. Both controllers determine ON/OFF stimulation of each muscle based on the crank angle and a pre-defined stimulation pattern which varies with the cadence to compensate for the delayed responses of the muscles. The stimulation intensity of the ON muscle is then computed based on the cadence error, PID gains, and a pre-defined Fuzzy logic rule.

\subsection{Neuromuscular Model}
\begin{figure}[h]
    \begin{center}
    \includegraphics[width=0.95\columnwidth]{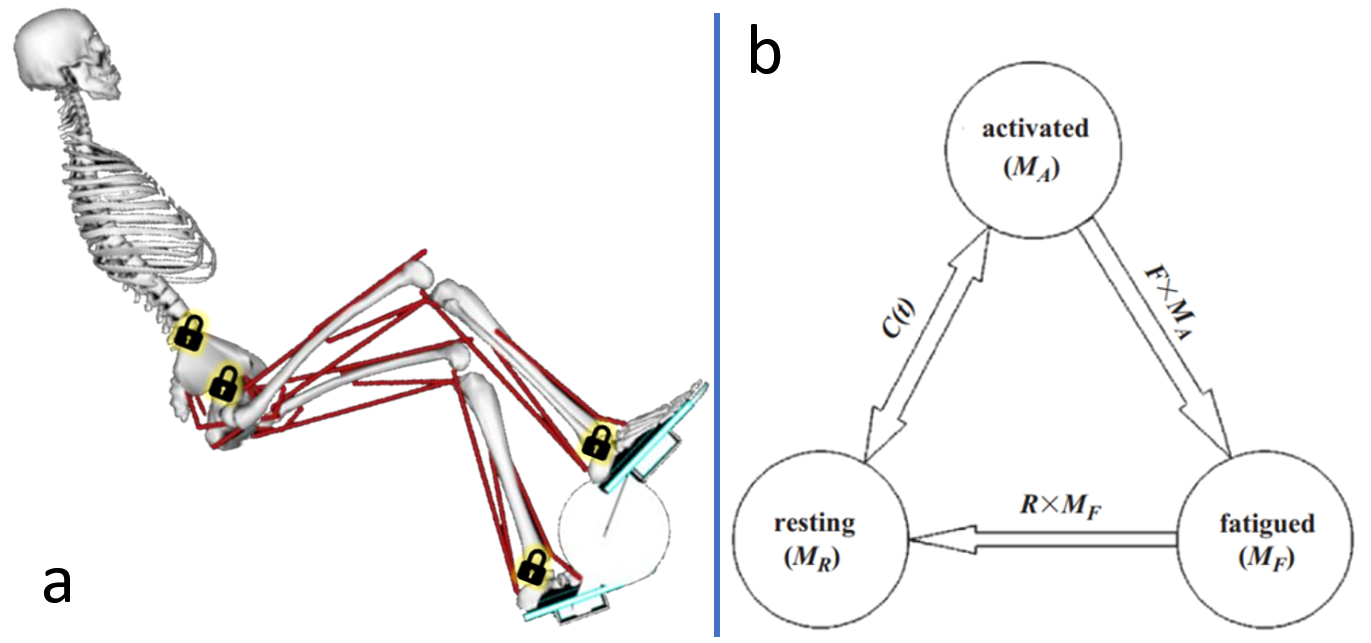}
    \captionsetup{margin=2mm, belowskip=-6mm, font=small}
    \caption{(a) the cycling musculoskeletal model. The locks indicate the locked angles. (b) Multi-state model of muscle fatigue.
    \label{fig:Figure23Natb}}
    \end{center}
\end{figure}

We trained and tested all controllers using a musculoskeletal model of a human skeleton with muscles and tendons using a crankset in a recumbent cycling position (Fig. \ref{fig:Figure23Natb} a). The model was built by De Sousa et al. \cite{DeSousa2016} in OpenSim, an open-source musculoskeletal simulation software. Our simulation includes the stimulation of the rectus femoris, the gluteus maximus, and the hamstrings on both legs. The stimulation of any other muscle is set to 0. The pelvis, lumbar and knee angles are fixed and remain unchanged during the simulation. The model includes a crankset which rotates without friction nor ground resistance.

\subsection{Muscle Fatigue Model}
Further to \cite{DeSousa2016} we include muscular fatigue in the musculoskeletal model as follows. Originally, the muscle force in Opensim is computed as $F^M = F_o^M(a\,f_L^M\,f_V^M + f_V^{PE})$, where $F^M$ is the muscle force; $F_o^M$ is the maximum isometric force;  $a$ is the muscle activation; $f_L^M$ is the active-force-length factor; $f_V^M$ is the force-velocity factor and $f_V^PE$ is the passive-force-length factor. We introduce a muscle fatigue factor $f_f^M$ to the active force part of the original equation as $F^M = F_o^M(a f_L^M f_V^M f_f^M + f_V^{PE})$, where $f_f^M$ has the values which are between $0$ (fully fatigued) and $1$.

The fatigue factor is computed based on the muscle fatigue model of Xia et al.\cite{Xia2008}. This muscle fatigue model considers a muscle as a group of muscle fibres which can be in one of the following states: resting ($M_R$), activated ($M_A$), and fatigued ($M_F$). The variables $M_R$, $M_A$, and $M_F$ are the fractions of muscle fibres being in each state. The transition of the fibres between each state is illustrated in Fig. \ref{fig:Figure23Natb} (b).

The transition rates between the states are determined by the coefficient $F$, fatigue rate, and $R$, recovery rate. The values of the coefficients can be obtained by fitting the model with experimental data. We used the values fitted from experiment data in \cite{Hainaut1989}. Note that, in the training, the fatigue and recovery rates are set to be 5 times of the original values to allow the agents to experience a wide range of fatigue level within a short training episode. $C(t)$ denotes the muscle activation–deactivation drive. The calculation of $C(t)$ is modified to include the relationship between stimulation ($s$) and muscle activation ($a$) to make the fatigue model compatible with Opensim as Eq. \ref{eq:ct}:

\begin{equation}
    C_{t} = 
\begin{cases}
    s - M_A & , \text{if } s \geq a\;\text{and}\;s-M_A \leq M_R \\
    M_R & , \text{if } s \geq a\;\text{and}\;s-M_A > M_R \\
    s - M_A & , \text{if } s \leq a
\end{cases}
\label{eq:ct}
\end{equation}

\comment{
The change of the fraction of the muscle fibres being in each state is computed as Equation \ref{eq:dMR}, \ref{eq:dMA}, and \ref{eq:dMF}:
\begin{equation}
    \frac{dMR}{dt} = -c(t) + R \times MF
\label{eq:dMR}
\end{equation}

\begin{equation}
    \frac{dMA}{dt} = c(t) - F \times MA
\label{eq:dMA}
\end{equation}

\begin{equation}
    \frac{dMF}{dt} = F \times MA - R \times MF
\label{eq:dMF}
\end{equation}
}

The variables $M_R$, $M_A$, and $M_F$ are then computed using first-order ordinary differential equations, the details of which can be found in \cite{Xia2008}. Finally, the fatigue factor $f_f^M$ is computed from the fraction of the muscle fibres that are not in fatigued state as $f_f^M = 1 - M_F$.

\section{Results}

\begin{figure}[htb]
\includegraphics[width=0.95\columnwidth]{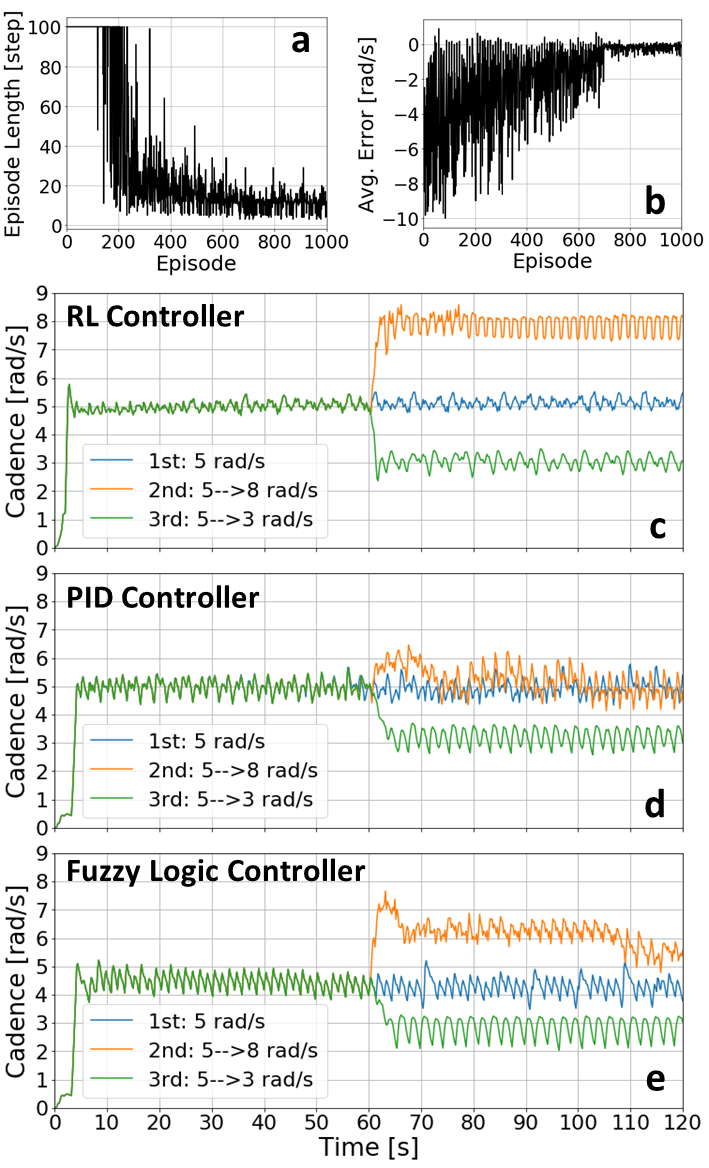}
\captionsetup{margin=2mm, belowskip=-5mm, font=small}
    \caption{(a) Learning curve of  Starter Agent. \label{fig:Figure45Nat_e0}. (b) Learning curve of  Tracker Agent. Performance for  (c) Deep RL, (d) PID and (e) Fuzzy Logic controllers for each of the 3 desired cadence cases. Note, PID and Fuzzy Logic struggle with muscle fatigue when the cadence target switches up and stays constant after 60s.}
\end{figure}

We create an RL controller to learn the stimulation strategy to cycle at the desired cadences. We trained the agents on a musculoskeletal model with fatigable muscles. The RL agents are trained for $1000$ episodes. The training of the Starter is monitored by the episode length (Fig. \ref{fig:Figure45Nat_e0} (a)). As the training progresses, the episode length decreases and the agent can reach the target cadence quicker. The training of the Tracker is monitored by the average cadence error (Fig. \ref{fig:Figure45Nat_e0} (b)). Here, as the training progresses, the error becomes closer to zero as the agent learns to control the cadence.

We compared the control performance of the RL controller with those of the PID and the Fuzzy Logic controllers. We test all controllers on three cases of desired cadence cases. All cases require the cadence of $5\,$rad/s in the first half of the sessions. In the second half, the requirements are $5$, $8$, and $3\,$rad/s in the first, second, and third cases, respectively. Fig. \ref{fig:Figure45Nat_e0} (c) shows the RL-controlled cadences along the test cycling sessions. The RL controller can produce the cadences that satisfy all three cases. The fluctuation is slightly high at $8\,$rad/s because the time-step size is slightly too large. Note that the RL controller is trained only once to track different random desired cadences specified at the beginning of each training episode. The values of the desired cadences are augmented into the state vectors observed by the Tracker.

The cadence results controlled by the PID and the Fuzzy Logic controllers are shown on Fig. \ref{fig:Figure45Nat_e0} (d) and (e), respectively.
Both controllers can produce the cadences that satisfy the 1\textsuperscript{st} and 3\textsuperscript{rd} cases. but fail to increase the cadence to $8\,$rad/s in the 2\textsuperscript{nd} case when the muscles are fatigued.

The control performances of the controllers are quantitatively evaluated in two aspects: tracking accuracy and response time. The tracking accuracy is evaluated using root mean square error (RMSE) computed from the second half of the sessions.
This calculation gives the RMSE when the desired cadences are $5$, $8$, and $3\,$ rad/s. Table \ref{tab:ResultTables} (A) shows the RMSEs of three controllers. The RMSEs of the RL controller are the lowest in all cases. This result means the RL controller has better tracking accuracy than the conventional controllers.

The response time is used to evaluate how quickly the controller can change the cadence to a new desired value. Here, we evaluate the response time at three cases shown in Table \ref{tab:ResultTables} (B). Among the 3 controllers, the RL one takes the shortest time in all cases, which means the RL controller has the best response time.
\vspace{-1 mm}
\begin{table}[h!]
    \begin{center}
    \includegraphics[width=\linewidth, trim=1 1 1 1, clip]{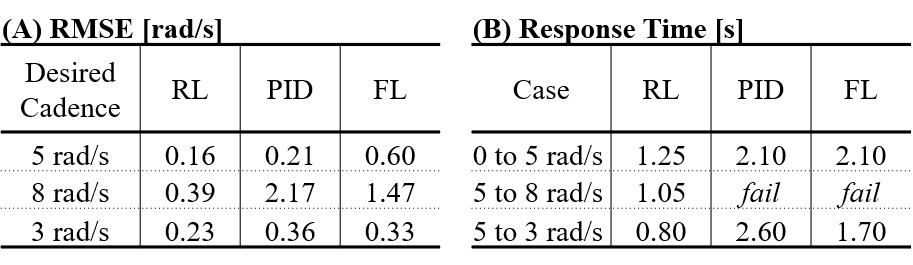}
    \captionsetup{margin=1mm, belowskip=-5mm, font=small}
    \caption{The tables showing RSME (A) and response time (B) of three controllers in three desired cadence pattern cases.
    \label{tab:ResultTables}}
    \end{center}
\end{table}

We participated in the FES Cycling race of the Cybathlon 2020 bionic Olympic games. We first modified the geometry of the Opensim model to match our paraplegic cyclist body (age 25, paralysed from the chest down). Then, we converted the learned control policy into a neurostimulation pattern which we uploaded into a commercial FES cycling neurostimulator and bicycle (Berkel Bikes, Sint-Michielsgestel, Netherlands). In the final, the athlete covered a distance of $1.2\,$km in 3 races and won the Silver medal with a time of $177\,$seconds in the international competition against other cyclists including a cyclist who used implanted electrodes.

\vspace{-2 mm}
\section{Conclusion \& Discussion}
We developed a Deep RL neurostimulation controller to control the electrical stimulation of 6 leg muscles to cycle at any desired cadences from start to steady-state. Trained in our neuromechanical simulation environment, our entire machine learned the system successfully and learned a control strategy whose quantified control accuracy and the quickness of response outperforms 20th century engineered controllers while also compensating for muscle fatigue.

There are, however, limitations of implementing our RL controller in real-world. Firstly, our RL controller, specifically the Tracker, needs to observe muscle fatigue level, which is not yet possible in real-world. Secondly, the amount of interaction data required for our RL controller to learn the policy is still too large to obtain from the interaction in real-world. To alleviate this issue, the RL controller can be pre-trained in simulation before being transferred to real humans. We conduct a simulation study in which a trained RL controller is transferred to another model with moderately different seat position. The preliminary result is that the RL controller can cycle on the new model immediately and requires around 10 minutes of additional interaction with the new model to produce good tracking performance. The transfer in real-world has yet to be investigated.

Regarding to the performance comparison between the RL and the conventional controllers, the RL controller here has privileged access to the fatigue information, allowing it to deal with the fatigue better than the conventional controllers. Nevertheless, the RL controller successfully learns to interpret and exploit the information without any pre-defined rule. This means it may be possible to integrate feedback signals that can provide  information on muscle fatigue such as electromyography (EMG) or mechanomyography (MMG) or both  \cite{Fara2013,xiloyannis2017gaussian} into a FES cycling system\cite{Woods2018}, controlled by an RL system. On the conventional side, there are some advanced methods such as HOSM \cite{Farhoud2014} that can maintain the cadence against the advancing fatigue in real-world. Compared to our method, a major difference is that the HOSM requires a pre-defined stimulation pattern (ON/OFF angles) while our method can learn the pattern by itself. Additionally, our method learns the pattern and the intensity control as a whole, allowing it to have refined stimulation for a specific situation, e.g., when the acceleration is needed. Our method also allows the controller to perform better over the time. This may has beneficial mental effects on the users.

We took our algorithms from the lab to the real-world evaluation by implementing our proof-of-principle controller on an actual FES bike for a paralysed cyclist to an international competitive sports event. The stimulation pattern discovered by our method combined with our cyclist's strategy on controlling the stimulation intensity allowed our team to achieve the second-best time in the competition. This real-world result on the specific task presents the potential and feasibility of machines to learn end-to-end neurostimulation for the restoration of complex function in the limbs of paralysed people.
\vspace{-2 mm}
\bibliographystyle{IEEEtran}
\bibliography{IEEEabrv}

\end{document}